\pdfoutput=1
\documentclass{svproc}

\usepackage{amssymb,amsmath,epsfig,latexsym,graphicx,bm,float}

\usepackage[hyphens]{url}
\usepackage[final=true, breaklinks=true]{hyperref}

\usepackage{epstopdf}

\begin{document}

\mainmatter              
\title{Machine Translation from Natural Language to Code using Long-Short Term Memory}
\titlerunning{Natural Language to Code}
\author{
K.M. Tahsin Hassan Rahit\inst{1} \inst{*} \and
Rashidul Hasan Nabil\inst{2} \and Md Hasibul Huq\inst{3}
}
\authorrunning{Tahsin et al.}
\institute{
Institute of Computer Science, Bangladesh Atomic Energy Commission, Dhaka, Bangladesh\\
*current - Department of Bio-chemistry \& Molecular Biology, University of Calgary, Calgary, Alberta, Canada\\
\email{kmtahsinhassan.rahit@ucalgary.ca}
\and
Department of Computer Science, American International University-Bangladesh, Dhaka, Bangladesh\\
Department of Computer Science \& Engineering, City University, Dhaka, Bangladesh\\
\email{merhnabil@gmail.com}
\and
Department of Computer Science and Software Engineering, Concordia University, Montreal, Quebec, Canada \\
\email{mdhasibul.huq@mail.concordia.ca}
}

\maketitle

\begin{abstract}
Making computer programming language more understandable and easy for the human is a longstanding problem. From assembly language to present day's object-oriented programming, concepts came to make programming easier so that a programmer can focus on the logic and the architecture rather than the code and language itself. To go a step further in this journey of removing human-computer language barrier, this paper proposes machine learning approach using Recurrent Neural Network(RNN) and Long-Short Term Memory(LSTM) to convert human language into programming language code. The programmer will write expressions for codes in layman's language, and the machine learning model will translate it to the targeted programming language. The proposed approach yields result with 74.40\% accuracy. This can be further improved by incorporating additional techniques, which are also discussed in this paper. 
\end{abstract}

\vspace{6mm}
\begin{keywords}
Text to code, machine learning, machine translation, NLP, RNN, LSTM
\end{keywords}

\section{Introduction}
Removing computer-human language barrier is an inevitable advancement researchers are thriving to achieve for decades. One of the stages of this advancement will be coding through natural human language instead of traditional programming language. On naturalness of computer programming  D. Knuth said, \textit{“Let us change our traditional attitude to the construction of programs: Instead of imagining that our main task is to instruct a computer what to do, let us concentrate rather on explaining to human beings what we want a computer to do.”}\cite{Knuth1984LiterateProgramming}. Unfortunately, learning programming language is still necessary to instruct it. Researchers and developers are working to overcome this human-machine language barrier. Multiple branches exists to solve this challenge (i.e. inter-conversion of different programming language to have universally connected programming languages). Automatic code generation through natural language is not a new concept in computer science studies. However, it is difficult to create such tool due to these following three reasons--
\begin{enumerate}
    \item Programming languages are diverse
    \item An individual person expresses logical statements differently than other
    \item Natural Language Processing (NLP) of programming statements is challenging since both human and programming language evolve over time

\end{enumerate}
In this paper, a neural approach to translate pseudo-code or algorithm like human language expression into programming language code is proposed.

\section{Problem Description}
Code repositories (i.e. Git, SVN) flourished in the last decade producing big data of code allowing data scientists to perform machine learning on these data. In 2017, Allamanis M \textit{et al.} published a survey in which they presented the state-of-the-art of the research areas where machine learning is changing the way programmers code during software engineering and development process \cite{Allamanis2017}. This paper discusses what are the restricting factors of developing such text-to-code conversion method and what problems need to be solved--

\subsection{Programming Language Diversity}
According to the sources, there are more than a thousand actively maintained programming languages, which signifies the diversity of these language\footnote{\url{https://en.m.wikipedia.org/wiki/List_of_programming_languages}} \footnote{\url{http://www.99-bottles-of-beer.net}}. These languages were created to achieve different purpose and use different syntaxes. Low-level languages such as assembly languages are easier to express in human language because of the low or no abstraction at all whereas high-level, or Object-Oriented Programing (OOP) languages are more diversified in syntax and expression, which is challenging to bring into a unified human language structure. Nonetheless, portability and transparency between different programming languages also remains a challenge and an open research area. George D. \textit{et al.} tried to overcome this problem through XML mapping \cite{GeorgeFCRIT2010}. They tried to convert codes from C++ to Java using XML mapping as an intermediate language. However, the authors encountered challenges to support different features of both languages. 

\subsection{Human Language Factor}
One of the motivations behind this paper is - as long as it is about programming, there is a finite and small set of expression which is used in human vocabulary. For instance, programmers express a \textit{for-loop} in a very few specific ways \cite{Mihalcea2006}. Variable declaration and value assignment expressions are also limited in nature. Although all codes are executable, human representation through text may not due to the semantic brittleness of code. Since high-level languages have a wide range of syntax, programmers use different linguistic expressions to explain those. For instance, small changes like swapping function arguments can significantly change the meaning of the code. Hence the challenge remains in processing human language to understand it properly which brings us to the next problem-

\subsection{NLP of statements}
Although there is a finite set of expressions for each programming statements, it is a challenge to extract information from the statements of the code accurately. Semantic analysis of linguistic expression plays an important role in this information extraction. For instance, in case of a loop, what is the initial value? What is the step value? When will the loop terminate?

Mihalcea R. \textit{et al.} has achieved a variable success rate of 70-80\% in producing code just from the problem statement expressed in human natural language \cite{Mihalcea2006}. They focused solely on the detection of step and loops in their research. Another research group from MIT, Lei \textit{et al.} use a semantic learning model for text to detect the inputs. The model produces a parser in C++ which can successfully parse more than 70\% of the textual description of input \cite{Lei2013}. The test dataset and model was initially tested and targeted against ACM-ICPC participants\' inputs which contains diverse and sometimes complex input instructions.

A recent survey from Allamanis M. \textit{et al.} presented the state-of-the-art on the area of naturalness of programming \cite{Allamanis2017}. A number of research works have been conducted on text-to-code or code-to-text area in recent years. In 2015, Oda \textit{et al.} proposed a way to translate each line of Python code into natural language pseudocode using Statistical Machine Learning Technique (SMT) framework \cite{Oda2015} was used. This translation framework was able to - it can successfully translate the code to natural language pseudo coded text in both English and Japanese. In the same year, Chris Q. \textit{et al.} mapped natural language with simple if-this-then-that logical rules \cite{Quirk2015}. Tihomir G. and Viktor K. developed an Integrated Development Environment (IDE) integrated code assistant tool \textit{anyCode} for Java which can search, import and call function just by typing desired functionality through text \cite{Gvero2015}. They have used model and mapping framework between function signatures and utilized resources like WordNet, Java Corpus, relational mapping to process text online and offline.

Recently in 2017, P. Yin and G. Neubig proposed a semantic parser which generates code through its neural model \cite{Yin2017}. They formulated a grammatical model which works as a skeleton for neural network training. The grammatical rules are defined based on the various generalized structure of the statements in the programming language.

\section{Proposed Methodology}
The use of machine learning techniques such as SMT proved to be at most 75\% successful in converting human text to executable code. \cite{Birch2008}. A programming language is just like a language with less vocabulary compared to a typical human language. For instance, the code vocabulary of the training dataset was 8814 (including variable, function, class names), whereas the English vocabulary to express the same code was 13659 in total. Here, programming language is considered just like another human language and widely used SMT techniques have been applied.

\subsection{Statistical Machine Translation}
SMT techniques are widely used in Natural Language Processing (NLP). SMT plays a significant role in translation from one language to another, especially in lexical and grammatical rule extraction. In SMT, bilingual grammatical structures are automatically formed by statistical approaches instead of explicitly providing a grammatical model. This reduces months and years of work which requires significant collaboration between bi-lingual linguistics. Here, a neural network based machine translation model is used to translate regular text into programming code.

\subsubsection{Data Preparation}
\begin{figure*}[t]
  \centering
  \includegraphics[width=\linewidth]{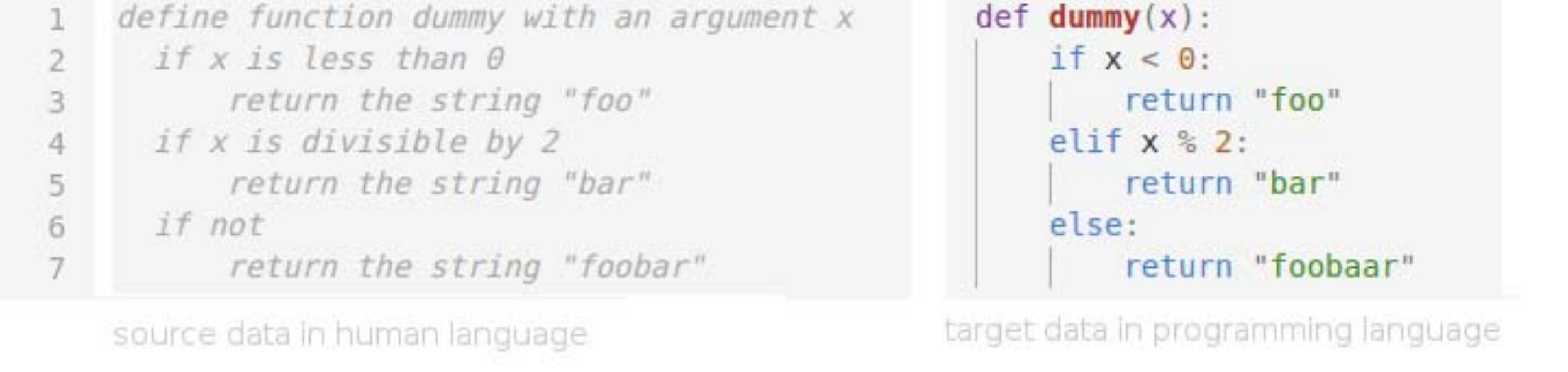}\\
  \caption{Text-Code bi-lingual corpus}\label{Fig:corpus}
\end{figure*}

SMT techniques require a parallel corpus in thr source and thr target language. A text-code parallel corpus similar to Fig. \ref{Fig:corpus} is used in training. This parallel corpus has 18805 aligned data in it \footnote{{Dataset: \url{https://ahclab.naist.jp/pseudogen/}} \cite{Oda2015}}. In source data, the expression of each line code is written in the English language. In target data, the code is written in Python programming language.

\begin{figure*}[t]
    \hspace{-3cm}
  \includegraphics[width=18cm, height=10cm]{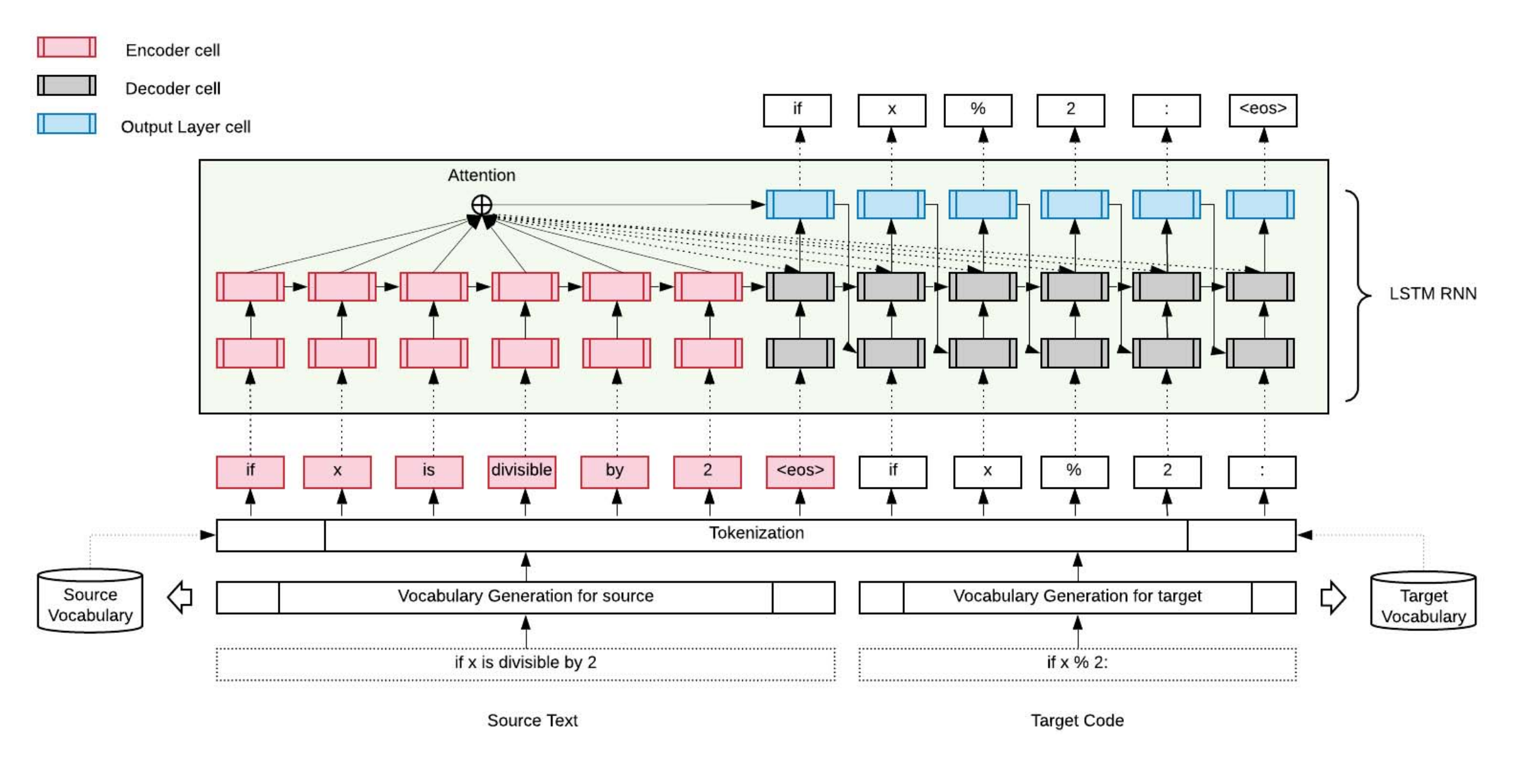}\\
  \caption{Neural training model architecture of Text-To-Code}\label{Fig:arch}
\end{figure*}
\subsubsection{Vocabulary Generation}
To train the neural model, the texts should be converted to a computational entity. To do that, two separate vocabulary files are created - one for the source texts and another for the code. Vocabulary generation is done by tokenization of words. Afterwards, the words are put into their contextual vector space using the popular \textit{word2vec} \cite{Mikolov2000} method to make the words computational. 

\subsubsection{Neural Model Training}
In order to train the translation model between text-to-code an open source Neural Machine Translation (NMT) - \textit{OpenNMT} implementation is utilized \cite{Klein2017OpenNMT:Translation}. \textit{PyTorch\footnote{\url{https://pytorch.org/}}} is used as Neural Network coding framework. For training, three types of Recurrent Neural Network (RNN) layers are used -- an encoder layer, a decoder layer and an output layer. These layers together form a LSTM model. LSTM is typically used in \textit{seq2seq} translation. 

In Fig. \ref{Fig:arch}, the neural model architecture is demonstrated. The diagram shows how it takes the source and target text as input and uses it for training. Vector representation of tokenized source and target text are fed into the model. Each token of the source text is passed into an encoder cell. Target text tokens are passed into a decoder cell. Encoder cells are part of the encoder RNN layer and decoder cells are part of the decoder RNN layer. End of the input sequence is marked by a \textit{$<$eos$>$} token. Upon getting the \textit{$<$eos$>$} token, the final cell state of encoder layer initiate the output layer sequence. At each target cell state, \textit{attention} is applied with the encoder RNN state and combined with the current hidden state to produce the prediction of next target token. This predictions are then fed back to the target RNN. \textit{Attention} mechanism helps us to overcome the fixed length restriction of encoder-decoder sequence and allows us to process variable length between input and output sequence. \textit{Attention} uses encoder state and pass it to the decoder cell to give particular attention to the start of an output layer sequence. The encoder uses an initial state to tell the decoder what it is supposed to generate. Effectively, the decoder learns to generate target tokens, conditioned on the input sequence. Sigmoidal optimization is used to optimize the prediction.

\section{Result Analysis}
\begin{figure}[H]
  \centering
  \includegraphics[width=\linewidth, height=5.5cm]{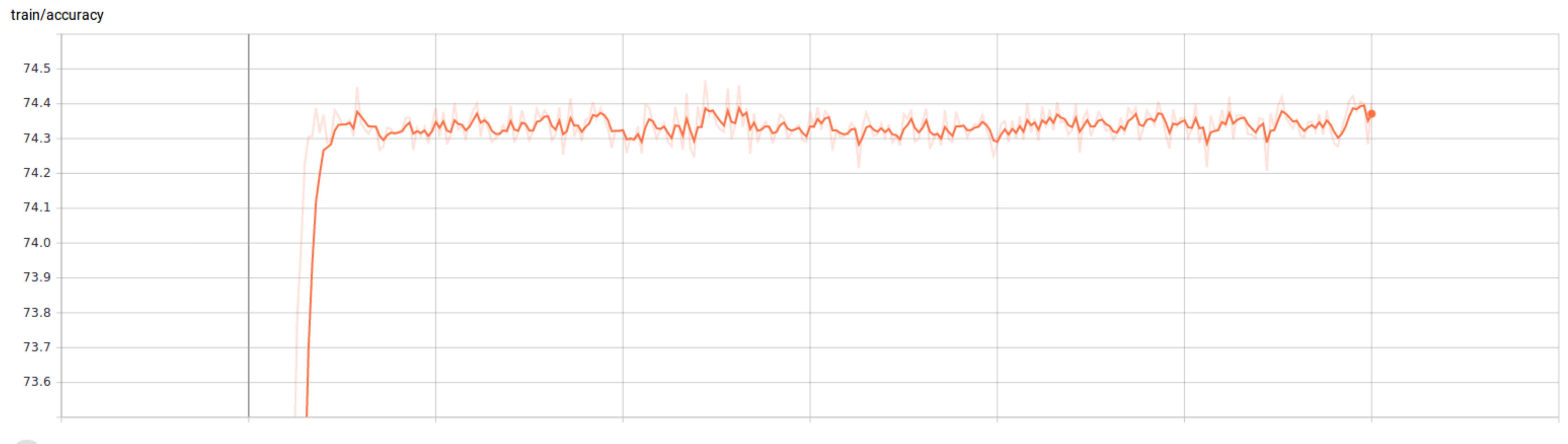}\\
  \caption{Accuracy gain in progress of training the RNN}\label{Fig:accuracy}
\end{figure}

Training parallel corpus had 18805 lines of annotated code in it. The training model is executed several times with different training parameters. During the final training process, 500 validation data is used to generate the recurrent neural model, which is 3\% of the training data. We run the training with epoch value of 10 with a batch size of 64. After finishing the training, the accuracy of the generated model using validation data from the source corpus was 74.40\% (Fig. \ref{Fig:accuracy}).

Although the generated code is incoherent and often predict wrong code token, this is expected because of the limited amount of training data. LSTM generally requires a more extensive set of data (100k+ in such scenario) to build a more accurate model. The incoherence can be resolved by incorporating coding syntax tree model in future. For instance-- 

\hspace{-0.4cm}\textit{"define the method tzname with 2 arguments: self and dt."}

\hspace{-0.3cm}is translated into--

\hspace{-0.3cm}\texttt{def \_\_init\_\_ ( self , regex ) :}.

\hspace{-0.3cm}The translator is successfully generating the whole codeline automatically but missing the noun part (parameter and function name) part of the syntax. 

\section{Conclusion \& Future Works}
The main advantage of translating to a programming language is - it has a concrete and strict lexical and grammatical structure which human languages lack. The aim of this paper was to make the text-to-code framework work for general purpose programming language, primarily Python. In later phase, phrase-based word embedding can be incorporated for improved vocabulary mapping. To get more accurate target code for each line, Abstract Syntax Tree(AST) can be beneficial. 

The contribution of this research is a machine learning model which can turn the human expression into coding expressions. This paper also discusses available methods which convert natural language to programming language successfully in fixed or tightly bounded linguistic paradigm. Approaching this problem using machine learning will give us the opportunity to explore the possibility of unified programming interface as well in the future. 

\section*{Acknowledgment}
We would like to thank Dr. Khandaker Tabin Hasan, Head of the Depertment of Computer Science, American International University-Bangladesh for his inspiration and encouragement in all of our research works. Also, thanks to Future Technology Conference - 2019 committee for partially supporting us to join the conference and one of our colleague - Faheem Abrar, Software Developer for his thorough review and comments on this research work and supporting us by providing fund.

\sloppy
\bibliography{mendeley_v2}
\bibliographystyle{splncs03}

\end{document}